# Remote Sensing and Machine Learning-Based Analysis of Land Use and Vegetation Change in Dhaka District, Bangladesh

[a]Muhammad Masud Tarek, [b]Md. Alamgir Hossain, [c]Md. Samiul Islam, [d]Muntasir Hasan Kanchan

[a, b, d]Department of Computer Science and Engineering, State University of Bangladesh, Dhaka, Bangladesh
[c]Department of Computer Science, American International University - Bangladesh, Dhaka, Bangladesh
[b, c, d]Skill Morph Research Lab., Skill Morph, Dhaka, Bangladesh

[a]E-mail: tarek@sub.edu.bd
[b]E-mail: alamgir.cse14.just@gmail.com, ORCID: https://orcid.org/0000-0001-5120-2911
[c]E-mail: samiulislam.cse@gmail.com
[d]E-mail: muntasir@sub.edu.bd

**Corresponding Author:** Md. Alamgir Hossain (Email: alamgir.cse14.just@gmail.com)

**Abstract**

Rapid urbanization in Dhaka District, Bangladesh has triggered substantial alterations in land use and environmental conditions, necessitating systematic monitoring for informed urban planning and ecological sustainability. This study employs remote sensing data and machine learning techniques to analyze spatiotemporal changes in land cover and vegetation dynamics between 2019 and 2024. High-resolution satellite imagery from Sentinel-2 MSI and Landsat 8 was utilized to classify land cover types and compute spectral indices including the Normalized Difference Vegetation Index (NDVI), Normalized Difference Built-up Index (NDBI), and Normalized Difference Water Index (NDWI). A supervised machine learning approach incorporating Decision Tree, K-Nearest Neighbors (KNN), and Random Forest classifiers was applied using labeled geospatial training points within Google Earth Engine. Accuracy assessments were conducted using confusion matrices and kappa statistics. Results indicate a 59.5% increase in urban built-up areas and a significant decline in vegetation (−8.46%) and water bodies (−7.77%) over the five-year period. Land conversion from vegetated and aquatic areas to urban infrastructure was identified as a dominant trend. Among the models, Random Forest demonstrated the highest classification accuracy. These findings underscore the growing environmental pressures driven by unregulated urban expansion in Dhaka. The study highlights the potential of remote sensing and machine learning tools in providing timely, actionable data to support sustainable urban development, land-use regulation, and ecosystem conservation policies.

***Keywords***: Land Cover Change Detection; Vegetation Change; Normalized Difference Vegetation Index; Sustainable Urban Planning; NDVI;

## 1. Introduction

### 1.1 Background and Motivation

Remote sensing has become an indispensable tool for analyzing Earth surface dynamics, particularly in urban and peri-urban landscapes undergoing rapid transformation. With its ability to provide continuous, synoptic, and multi-temporal observations, satellite-based remote sensing facilitates comprehensive environmental monitoring, urban planning, and resource management. Among its analytical capabilities, spectral indices such as the Normalized Difference Vegetation Index (NDVI) play a pivotal role in evaluating vegetation health and density. NDVI, derived from the red and near-infrared bands, is widely used for detecting ecological changes and assessing biomass distribution [1], [2].

The fusion of remote sensing with machine learning (ML) algorithms, particularly ensemble models like Random Forest, has further advanced the accuracy and automation of land use and land cover (LULC) classification. These advancements are particularly relevant for urban environments experiencing fast-paced, unplanned growth. Dhaka, the capital of Bangladesh, is one of the world's most densely populated and rapidly urbanizing megacities [3], [4]. It is facing intense pressure on land resources due to population growth, infrastructure expansion, and informal settlement development. Over the last two decades, studies have documented the encroachment of wetlands, conversion of vegetated and agricultural lands, and the expansion of impervious surfaces. These transitions threaten ecological balance, increase flood risks, and degrade environmental quality, warranting detailed, high-resolution monitoring to guide sustainable development [5].

### 1.2 Research Gaps

Although numerous studies have assessed land cover dynamics in Bangladesh, most are conducted at a national or regional scale, often failing to capture localized patterns specific to Dhaka District. In addition, many rely solely on visual interpretation or conventional classification techniques, lacking integration of spectral indices (NDVI, NDBI, NDWI) with modern machine learning frameworks. Few have leveraged scalable cloud computing tools like Google Earth Engine (GEE) for efficient processing of large satellite datasets. Moreover, comparative evaluation of different ML classifiers under Dhaka's unique urban morphology remains underexplored, limiting methodological insights for similar urban environments in the Global South.

### 1.3 Research Objectives

This study aims to bridge these gaps through a multi-faceted geospatial analysis of Dhaka District using open-access Sentinel-2 imagery and Google Earth Engine. The specific objectives are:

- To monitor land use and land cover changes in Dhaka District over multiple years using high-resolution satellite data.
- To implement and compare the performance of machine learning classifiers for accurate LULC classification.
- To evaluate the changes in vegetation, urban infrastructure, and water bodies using spectral indices (NDVI, NDBI, NDWI) and compare their effectiveness with ML-based classifications.
- To analyze spatiotemporal trends in land transformation and assess their environmental and policy implications for urban sustainability.

### 1.4 Research Questions

This study is guided by the following research questions:

**RQ1:** What are the major land cover classes in Dhaka District, and how have they changed over recent years?

**RQ2:** How effective are machine learning models in classifying land cover types using remote sensing data?

**RQ3:** How do spectral index-based methods (NDVI, NDBI, NDWI) compare with machine learning classifications in detecting vegetation, water bodies, and built-up areas?

**RQ4:** What are the spatiotemporal trends in vegetation loss, urban expansion, and water body reduction in Dhaka District?

**RQ4:** How can the findings inform sustainable urban planning and environmental management policies in rapidly urbanizing regions like Dhaka?

Following this introduction, Section 2 provides a comprehensive review of the existing literature and background on remote sensing, land classification techniques, and NDVI, NDWI, and NDBI. Section 3 details the methodology used in this study, including data collection, preprocessing, and analysis techniques. Section 4 presents the data analysis, including the implementation of land classification and normalized difference indices analysis, and the results of the study, including implications and limitations. Finally, Section 5 concludes the paper with a summary of findings and recommendations for future research.

## 2. Literature Review and Theoretical Background

Remote sensing technology involves acquiring information about the Earth's surface without physical contact, primarily through the use of satellite or aerial sensors. This technology has become an essential tool for environmental monitoring, urban planning, agriculture, forestry, and disaster management. Remote sensing provides a synoptic view and repeated coverage, making it ideal for monitoring changes over time. Key advantages include the ability to collect data over inaccessible areas, the availability of historical datasets, and the capability to cover large areas efficiently [6], [7]. Remote sensing is a very wide field, so here we only focused on satellite imagery, particularly the public domain satellite imagery. The first thing we want to know is, how satellites see the world and how they are processed [8].

Passive remote sensing: Satellite sensors sense the amount of reflected sunlight from the earth surface. The sun's light is actually a form of electromagnetic radiation that consists of radiation in many different wavelengths and our eyes can sense only a small portion of radiation in visible spectrum band. Landsat8 launched in 2013 which was developed as a collaboration between NASA and the U.S. Geological Survey (USGS) [9]. The satellite payload includes two sensors: The Operational Land Imager (OLI), which provides nine spectral bands including a panchromatic band, and the Thermal Infrared Sensor (TIRS), which offers two thermal spectral bands. As illustrated in ***Figure 1***, passive remote sensing relies on natural energy, typically solar radiation, which is reflected by the Earth's surface and captured by satellite sensors.

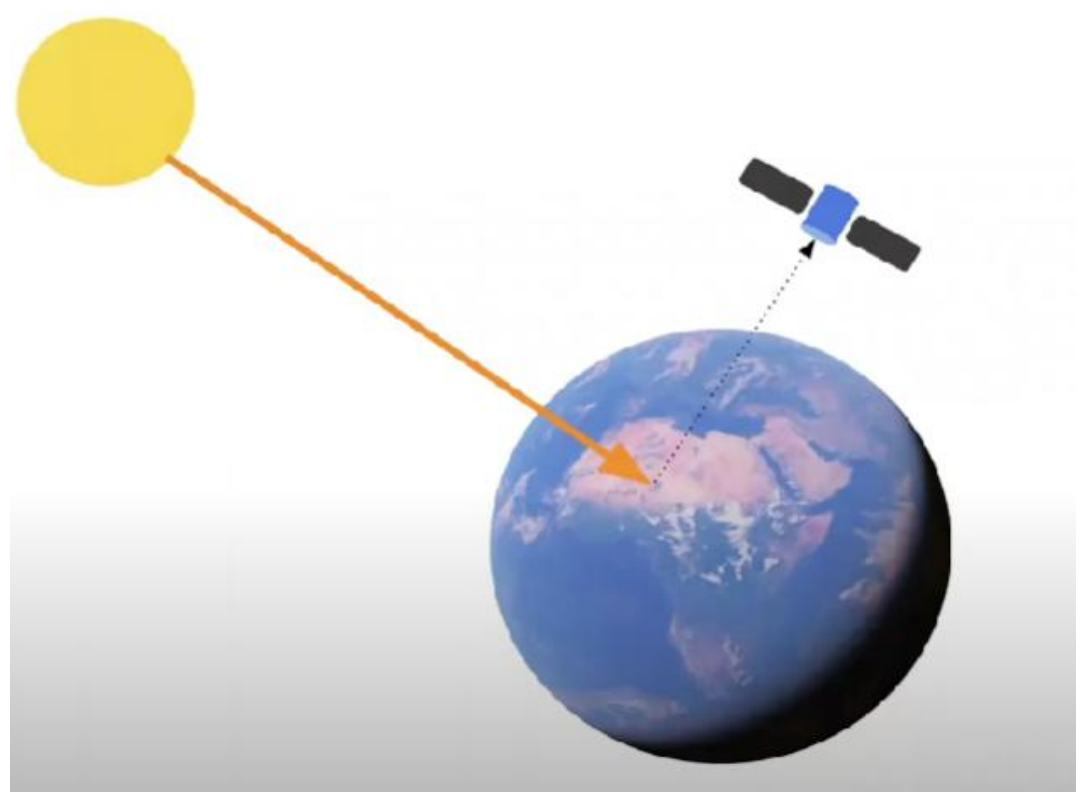

**Figure 1** Illustration of passive remote sensing, where solar radiation is reflected from the Earth's surface and detected by a satellite sensor without actively emitting any signal.

Landsat 8 captures imagery through multiple spectral bands ranging from visible to thermal wavelengths, each with varying spatial resolutions suited for different types of land surface analysis shown in ***Table 1***.

**Table 1** Spectral bands, wavelength ranges, and spatial resolutions of Landsat 8 OLI and TIRS sensors.

| Band Number | Description | Wavelength | Resolution |
|---|---|---|---|
| Band 1 | Coastal / Aerosol | 0.433 to 0.453 µm | 30 meters |
| Band 2 | Visible blue | 0.450 to 0.515 µm | 30 meters |
| Band 3 | Visible green | 0.525 to 0.600 µm | 30 meters |
| Band 4 | Visible red | 0.630 to 0.680 µm | 30 meters |
| Band 5 | Near-infrared | 0.845 to 0.885 µm | 30 meters |
| Band 6 | Short wavelength infrared | 1.56 to 1.66 µm | 30 meters |
| Band 7 | Short wavelength infrared | 2.10 to 2.30 µm | 60 meters |
| Band 8 | Panchromatic | 0.50 to 0.68 µm | 15 meters |
| Band 9 | Cirrus | 1.36 to 1.39 µm | 30 meters |
| Band 10 | Long wavelength infrared | 10.3 to 11.3 µm | 100 meters |
| Band 11 | Long wavelength infrared | 11.5 to 12.5 µm | 100 meters |

The European Space Agency (ESA), under the Copernicus Earth observation program, launched the Sentinel satellite series (Sentinel-1, -2, -3, and -5P) between 2014 and 2018 to support applications in land monitoring, marine environments, atmospheric studies, emergency response, security, and climate change [10]. Sentinel-2 provides 13 spectral bands with spatial resolutions ranging from 10 to 60 meters, including high-resolution visible and near-infrared bands (B2–B8) and lower-resolution bands for atmospheric correction and cirrus detection. Unlike Landsat 8, which uses passive optical sensors, Sentinel-1 employs active microwave sensors, making it effective for all-weather imaging, particularly flood detection. Sentinel-5P focuses on atmospheric monitoring using ultraviolet and infrared signals. The spectral configuration and atmospheric transmission characteristics of Landsat sensors are illustrated in ***Figure 2***.

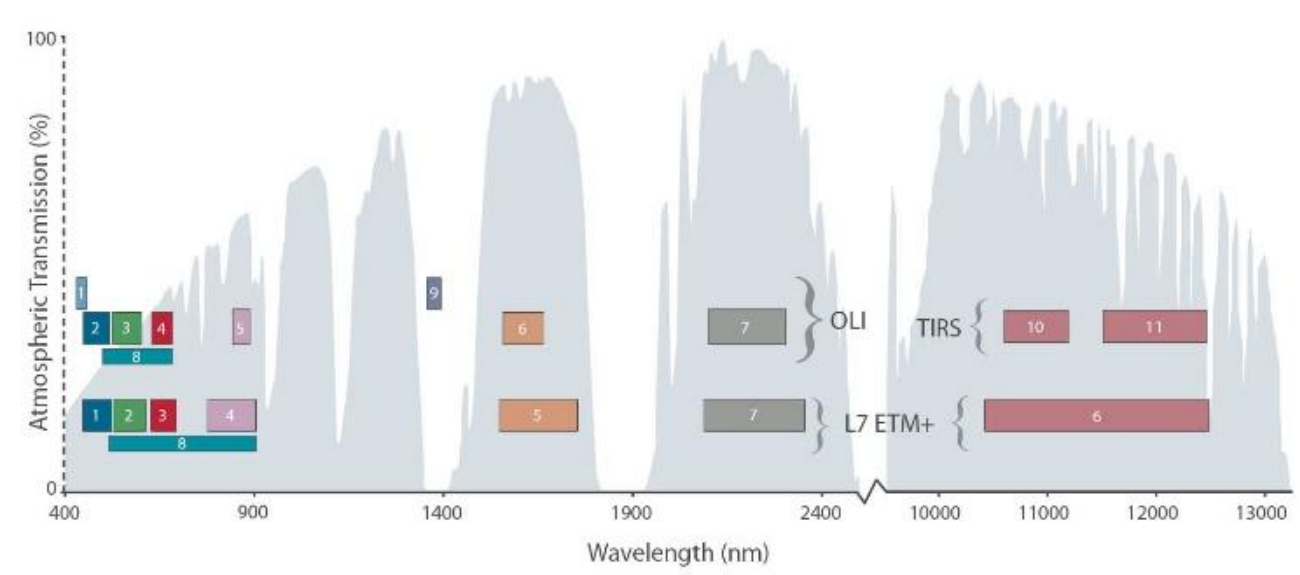

**Figure 2** Spectral band coverage and atmospheric transmission characteristics of Landsat 8's OLI and TIRS sensors.

Satellite data is processed through a series of standardized levels. Level 0 (L0) represents raw, unprocessed data directly collected by the satellite sensors, which is not suitable for application without calibration [11]. Level 1 (L1) data is corrected for radiometric (sensor calibration) and geometric distortions caused by satellite movement. This processed data, known as Top of Atmosphere (ToA) products, is ready for general analytical use. Level 2 (L2) involves atmospheric correction of L1 data to estimate surface reflectance or temperature, providing more accurate information about the Earth's surface. Level 3 (L3) converts scene-based data into a uniform grid format, making it suitable for integration into distributed systems and large-scale analysis.

The spatial resolution of a satellite image defines the size of the smallest object that can be detected, essentially the area each pixel represents on the ground [12]. Temporal resolution refers to the frequency at which a satellite revisits the same location. For instance, MODIS offers near-daily coverage using its AQUA and Terra satellites, while Landsat 8 revisits every 16 days, and Sentinel-2 provides a revisit interval of 10 days per satellite.

Spectral resolution indicates the number and width of spectral bands recorded in a scene. Multispectral sensors typically include 3 to 10 bands, whereas hyperspectral sensors capture hundreds of narrow bands, enabling the detection of specific vegetation species or material types.

Radiometric resolution defines the sensitivity of a sensor to detect subtle differences in intensity. Landsat 8 offers a 16-bit radiometric resolution, while Sentinel-2 and MODIS operate with 12-bit resolution. Another crucial characteristic is swath width, which refers to the width of the Earth's surface that the satellite can image in a single pass [13].

For visual interpretation, different band combinations are used to enhance features. Natural color images use the red, green, and blue (RGB) bands to resemble what the human eye perceives. False color infrared composites (using near-infrared, red, and green) highlight healthy vegetation in red tones. False color urban combinations (SWIR2, SWIR1, and red) display built-up areas in green. Agricultural composites (SWIR1, NIR, and blue) depict croplands in light green and distinguish bare soil in yellow hues.

Spectral indices and band ratio techniques are widely used in remote sensing for automated detection and analysis of land cover types such as vegetation, water bodies, and urban areas. These methods rely on the unique reflectance characteristics of different surfaces across spectral bands. For instance, vegetation exhibits strong reflectance in the near-infrared (NIR) band and low reflectance in the red band. This contrast forms the basis of the Normalized Difference Vegetation Index (NDVI), calculated as (NIR – Red) / (NIR + Red), with values ranging from –1 to +1. Higher values indicate denser, healthier vegetation.

To further account for soil background influence in sparse vegetation areas, the Soil-Adjusted Vegetation Index (SAVI) is applied, which introduces a correction factor. Similarly, water bodies show high reflectance in the green band and very low in NIR, which enables the Normalized Difference Water Index. An improved version, the Modified NDWI (MNDWI), replaces NIR with SWIR1 for better separation of water and built-up surfaces.

Built-up and bare soil areas typically reflect more strongly in SWIR than NIR. This forms the basis of the Normalized Difference Built-Up Index (NDBI). For Landsat 8 data, this corresponds to: NDBI = (Band 6 – Band 5) / (Band 6 + Band 5). To isolate built-up areas more precisely, the Built-Up Index (BU) is sometimes computed as the difference between NDBI and NDVI: BU = NDBI – NDVI.

These indices are computationally efficient and can be rapidly derived from satellite imagery. In contrast, land cover classification using machine learning methods offers greater accuracy but requires complex workflows and extensive training data. The reflectance behavior that underpins these indices across various land surfaces, such as vegetation, water, and soil is illustrated in ***Figure 3***.

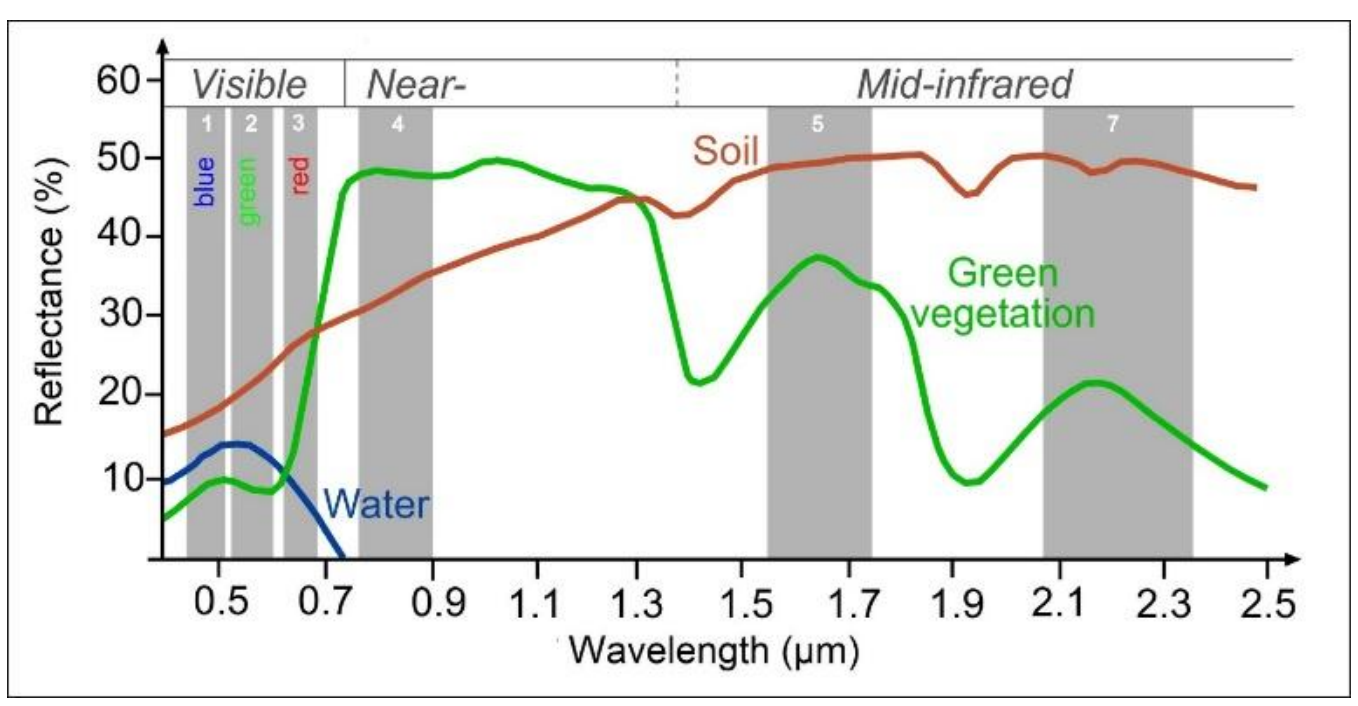

**Figure 3** Spectral Reflectance Curve. Normalized Differences Indices utilize the reflection properties of different bands over different types of land cover (*[14]*)

Satellite imagery is available through various data access models. Open data refers to freely available imagery, typically of medium to low resolution, suitable for most academic and governmental purposes [15]. Lagged data becomes available after a short processing delay, while commercial data offering very high resolution and more frequent revisits is accessible through paid licenses.

Processing latency also varies depending on data type and processing level. Real-time and near real-time data are typically available within a few hours (ranging from 0.5 to 6 hours), though they may have limited corrections and lower resolution. Standard products are released after full processing with delays ranging from one to sixteen days. Additionally, reprocessed datasets may become available when new algorithms or calibration techniques improve upon previously released data, enhancing accuracy and usability for long-term studies.

Google Earth Engine (GEE) is a cloud-based geospatial analysis platform designed to support large-scale environmental and remote sensing research. One of its core features is access to a massive geospatial data catalog, which includes pre-processed datasets such as Landsat, Sentinel imagery, climate records, elevation models, and socio-economic data. This extensive catalog reduces preprocessing time and enhances analytical efficiency. GEE also provides scalable cloud computing infrastructure through Google's cloud services, enabling users to process petabytes of data without relying on high-performance local machines. This capability makes GEE particularly effective for conducting global or regional studies, including deforestation tracking, flood mapping, and climate change analysis [16], [17].

Another key component of GEE is its scripting and API support, which allows users to develop advanced spatial analyses using JavaScript and Python. These APIs support complex operations such as image classification, temporal change detection, and machine learning integration. Additionally, GEE includes interactive visualization tools that allow users to overlay geospatial datasets, explore trends over time, and create dynamic, shareable maps and dashboards [18]. As illustrated in ***Figure 4***, these integrated features make

Google Earth Engine an essential tool for researchers, policymakers, and developers working in environmental monitoring, disaster response, and land use planning.

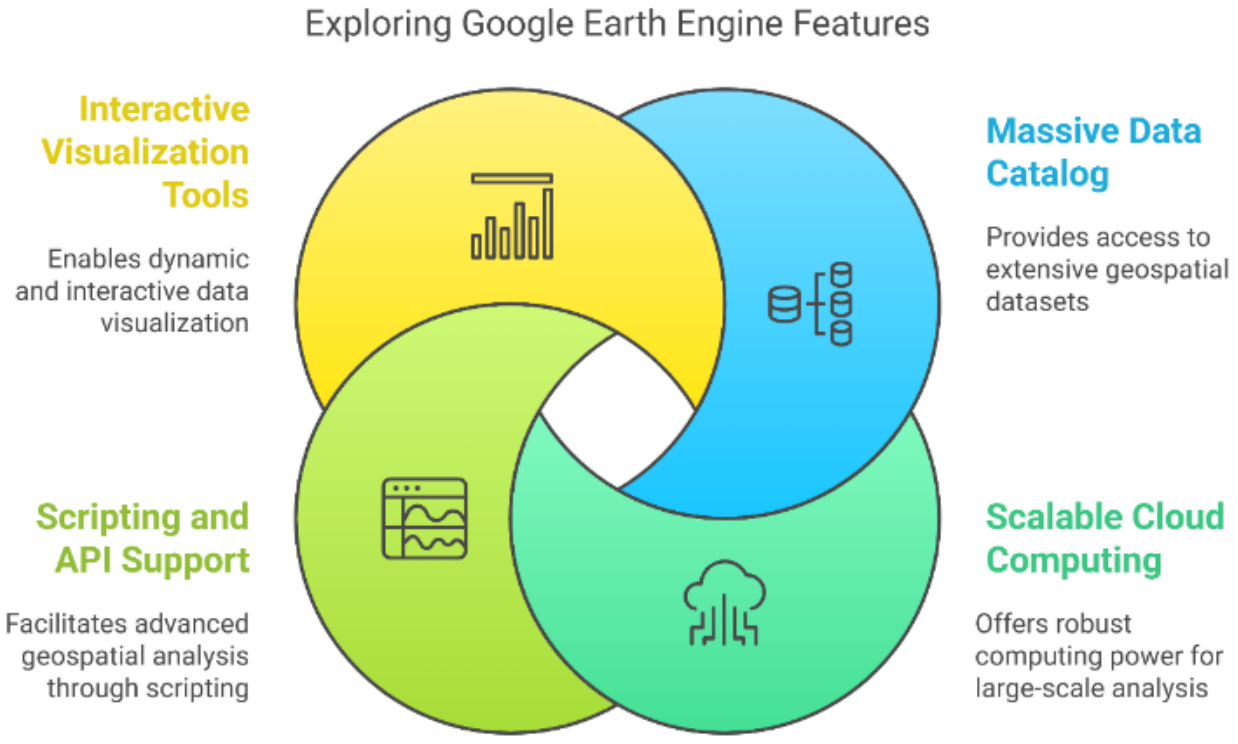

**Figure 4** Key features of Google Earth Engine, including its massive geospatial data catalog, scalable cloud computing infrastructure, interactive visualization tools, and scripting/API support for advanced geospatial analysis.

Land classification involves categorizing the Earth's surface into various land cover types such as urban areas, agricultural fields, forests, water bodies, and barren lands. Remote sensing has enabled efficient and scalable land classification by analyzing spectral signatures from satellite imagery. Several techniques have been developed for this purpose.

Traditional pixel-based classification methods are limited in handling mixed pixels and spectral overlap. Machine learning (ML) algorithms such as Random Forest (RF), Decision Trees (DT), and K-Nearest Neighbors (KNN) have proven effective for supervised land cover classification. They leverage statistical learning to recognize complex patterns and are particularly useful when working with multi-spectral satellite data. RF, for example, aggregates decisions from multiple decision trees and has shown high classification accuracy and generalizability in heterogeneous urban environments. Recent advancements in cloud-based processing platforms like Google Earth Engine (GEE) have further enabled scalable ML-based classification with minimal computing overhead [19], [20].

In contrast, unsupervised classification does not require prior knowledge of land cover types. Instead, it clusters pixels based on spectral similarity, with algorithms like K-means and ISODATA being commonly used. This approach is useful for exploratory analysis when training data is unavailable.

Another advanced method is Object-Based Image Analysis (OBIA), which segments images into spatially coherent objects using both spectral and spatial information. This technique is particularly effective in heterogeneous landscapes, such as urban environments, where traditional pixel-based methods often struggle with mixed pixels [21].

The Normalized Difference Vegetation Index (NDVI) is a widely recognized metric used to assess vegetation health, density, and coverage using satellite data. It is calculated using the near-infrared (NIR) and red (RED) bands, which correspond to Bands 8 and 4 in Sentinel-2 imagery [22].

NDVI values range from -1 to +1, with higher values indicating dense, healthy vegetation, and lower or negative values indicating water, barren land, or impervious surfaces. NDVI has been extensively applied in various domains including drought monitoring, deforestation analysis, crop health assessment, and vegetation phenology studies [23].

Remote sensing and NDVI-based methods have proven useful in understanding urban environmental dynamics, despite challenges such as spectral mixing and heterogeneous land cover types.

One prominent application is in the study of Urban Heat Islands (UHIs), where NDVI has shown a strong inverse correlation with land surface temperature. Areas with greater vegetation cover tend to exhibit lower temperatures, mitigating UHI effects [24].

Remote sensing is also critical in urban expansion monitoring, where temporal NDVI analysis reveals the loss of green spaces and vegetation due to built-up area growth. NDVI change detection facilitates a clearer understanding of spatial trends in land use transformation [25].

Additionally, environmental impact assessments often incorporate NDVI and remote sensing datasets to evaluate changes in ecosystem health, land cover alterations, and the degradation of vegetative surfaces caused by rapid urbanization.

Several studies have explored the environmental transformation of Dhaka District, highlighting the significant impacts of rapid urban growth. Research on land use change has documented the steady conversion of agricultural lands and peri-urban areas into densely built-up zones, accompanied by a marked decrease in vegetated cover and open spaces [26], [27].

Remote sensing has also been instrumental in flood risk mapping, where satellite-derived data has helped identify vulnerable zones and emphasized the ecological importance of preserving wetlands and natural drainage channels in Dhaka's urban planning [28]. Moreover, satellite-based air quality monitoring has linked increasing levels of atmospheric pollutants to expanding industrial and vehicular activity within the city, reflecting the broader consequences of uncontrolled urbanization [29].

While these studies provide critical insights, there remains a significant gap in integrating land classification techniques with NDVI-based change detection to create a comprehensive view of Dhaka's environmental trajectory. Expanding the analytical framework to include additional indices such as the Soil

Adjusted Vegetation Index (SAVI) and the Modified Normalized Difference Water Index (MNDWI) could further enhance the reliability and interpretability of findings.

## 3. Methodology

*3.1 Study Area*

Dhaka District, located in central Bangladesh, is characterized by its rapid urbanization and dense population. The district encompasses both urban and peri-urban areas, offering a diverse range of land cover types, including residential, commercial, agricultural, and water bodies. This diversity makes Dhaka an ideal study area for land classification and visual spectral analysis (NDVI, NDBI, NDWI).

*3.2 Data Collection*

For this study, satellite imagery from the Sentinel-2 MSI (Multi-Spectral Instrument) was utilized. Sentinel-2 MSI provides high-resolution images with multiple spectral bands, making it suitable for land classification and normalized difference Indices calculations. The administrative boundary data of Dhaka district was collected from The Food and Agriculture Organization of the United Nations provided data set. The specific data used in this research include:

- Multispectral Images: Acquired from the Sentinel-2 MSI, covering various wavelengths necessary for land cover classification and normalized difference indices (NDVI, NDWI, NDBI) computation.
- Temporal Data: Images from different time periods were collected to analyze changes in land cover and vegetation over time.
- Ancillary Data: Additional data, such as digital elevation models (DEMs) and land use maps, were used to support the classification process.
- Administrative boundary of Dhaka District of Bangladesh was collected from The Food and Agriculture Organization of the United Nations (FAOUN) 2015 Dataset, satellite image scale at 500 meters per pixel.

*3.3 Preprocessing*

To ensure the accuracy, consistency, and reliability of satellite imagery, several preprocessing steps were undertaken. Radiometric calibration was applied to correct sensor-related anomalies and to maintain consistent pixel value interpretation across different scenes. Geometric correction was performed to align all imagery to a common spatial reference system, enabling accurate overlay and spatial analysis. Cloud masking techniques were employed to remove cloud-contaminated pixels, which could otherwise result in misclassification during land cover analysis. Finally, image enhancement methods were used to improve visual clarity and interpretability, facilitating better feature extraction and analysis. These preprocessing procedures are critical in remote sensing workflows and are illustrated in ***Figure 5***, which outlines the key objectives of each step.

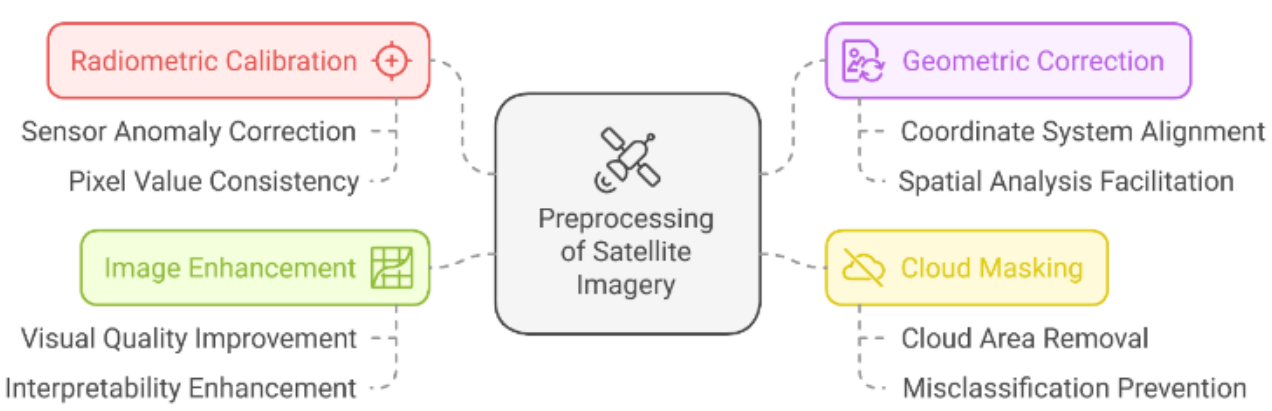

**Figure 5** Main preprocessing steps for satellite imagery to ensure accuracy and clarity.

*3.4 Cloud Removal Approach*

Cloud removal is a critical step in ensuring the quality of satellite imagery for analysis. One common approach is cloud masking and filtering, where Sentinel-2's Scene Classification Maps (SCL) and cloud probability layers are used to identify cloud-affected areas. Additionally, the Fmask (Function of Mask) algorithm is widely employed to detect clouds, shadows, and snow for exclusion from further processing. Temporal compositing and multi-date image mosaicking involve combining multiple images from different dates to replace cloud-covered regions with cloud-free pixels, often using techniques like Maximum Value Composite (MVC) or Median Composite. When cloud coverage is low (typically between 10–30%), cloud shadow interpolation and gap filling methods, such as spatial interpolation or deep learning models, can be applied to reconstruct missing data. Lastly, a harmonized time-series approach uses sequential imagery to selectively extract cloud-free pixels, ensuring the generation of a consistent and complete dataset for land cover classification and change detection.

*3.5 Time Period Selection for Image Data*

The selection of an appropriate time period for satellite image processing is crucial and largely depends on the specific application and the frequency of cloud cover in the study region. In areas with persistent cloudiness, seasonal or annual composite approaches are often adopted to reduce data gaps and ensure reliable surface observation. For targeted analyses such as monitoring peak vegetation growth multi-temporal analysis can be employed to compensate for cloud-obscured scenes by incorporating nearby cloud-free observations, thereby enhancing data continuity and accuracy.

*3.6 Land Classification Methods*

The land classification in this study was conducted using supervised classification techniques, which involve training the model with labeled data to categorize satellite imagery into distinct land cover types. Training data was gathered through field surveys and existing land use maps, providing reliable references for the classification process. Three machine learning algorithms were employed: Decision Tree (DT), a non-parametric algorithm that uses a tree-like structure to classify pixels based on their attributes; K-Nearest Neighbor (KNN), which classifies data points based on the majority class among their nearest neighbors; and Random Forest, an ensemble method that builds multiple decision trees and aggregates their outputs for improved accuracy and robustness.

To evaluate classification performance, accuracy assessment was carried out using confusion matrices and kappa statistics, which measure the agreement between predicted and actual land cover types. Additionally, a pretrained land cover classifier developed by the Food and Agriculture Organization of the United Nations (FAOUN) was used for comparison. This external benchmark helped validate the performance of the developed models and ensured the reliability of classification outcomes for the Dhaka District.

### *3.7 Indices Calculation and Change Detection*

Spectral indices are widely used in remote sensing for the automatic detection of urban areas, vegetation cover, and water bodies. These indices utilize the distinct reflectance properties of different land surface types across specific spectral bands to highlight particular features. The Normalized Difference Vegetation Index (NDVI) is one of the most common vegetation indices, calculated using:

$$NDVI = \frac{NIR - Red}{NIR + Red} \tag{1}$$

Where NIR and Red represent the near-infrared and red spectral bands, respectively. To reduce the influence of soil brightness, especially in areas with sparse vegetation, the Soil Adjusted Vegetation Index (SAVI) is used, defined as:

$$SAVI = \frac{1.5 * (NIR - Red)}{NIR + Red + 0.5} \tag{2}$$

For water detection, the Normalized Difference Water Index (NDWI) is applied:

$$NDWI = \frac{Green - NIR}{Green + NIR} \tag{3}$$

A modified version, the Modified NDWI (MNDWI), replaces the NIR band with SWIR1 for improved detection in urban settings:

$$MNDWI = \frac{Green - SWIR1}{Green + SWIR1} \tag{4}$$

To detect built-up areas, the Normalized Difference Built-up Index (NDBI) is used, which contrasts the SWIR1 and NIR bands:

$$NDBI = \frac{SWIR1 - NIR}{SWIR1 + NIR} \tag{5}$$

These indices were calculated for different years and compared to detect land cover changes in vegetation, water bodies, and urban areas across the study period.

Values were computed for multiple time periods to analyze vegetation dynamics and other land cover changes. The Normalized Difference Vegetation Index (NDVI), Modified Normalized Difference Water Index (MNDWI), and Normalized Difference Built-up Index (NDBI) were calculated using specific Sentinel-2 bands namely, Green (B3), Red (B4), Near Infrared (B8), and SWIR1 (B11) as shown in ***Table 2***. The change detection process included three main steps. First, a temporal comparison was conducted by evaluating index values across different years to identify significant changes in vegetation, water bodies, and built-up areas. Second, the machine learning (ML) classification results were compared with the index-derived land cover classifications to assess consistency and accuracy. Lastly, spatial analysis using Geographic Information System (GIS) tools enabled the visualization and interpretation of land cover transitions across the Dhaka District.

**Table 2** Selected Sentinel-2 bands used for NDVI, MNDWI, and NDBI calculations.

| Band | Wavelength names |
|---|---|
| B3 | Green |
| B4 | Red |
| B8 | NIR (Near Infrared) |
| B11 | SWIR1 (Short Wave Infrared channel 1) |

### 3.8 Tools and Software

This study utilized several tools to conduct preprocessing, classification, and analytical tasks efficiently. Google Earth Engine (GEE) served as the primary platform for satellite image preprocessing, land classification, spectral index calculation, and change detection analysis. Its cloud-based infrastructure enabled scalable and rapid processing of large geospatial datasets. In addition, JavaScript was used within the GEE environment to script data workflows and implement machine learning algorithms for land cover classification.

By adopting these tools and following the described methodology, the study delivers a comprehensive assessment of land cover transformations and vegetation dynamics in Dhaka District, offering critical insights to support sustainable urban planning and environmental decision-making.

## 4. Experimental Results Analysis

This section presents the experimental analysis conducted to evaluate land cover classification and NDVI-based vegetation change detection across Dhaka District. The analysis was performed using Google Earth Engine (GEE), a cloud-based geospatial processing platform that provides access to a comprehensive repository of satellite imagery, including Sentinel-2 and Landsat datasets. GEE’s built-in libraries and computational infrastructure support the efficient execution of remote sensing workflows at scale. JavaScript was utilized within the GEE code editor to implement core processes such as image preprocessing, spectral index calculation, and supervised classification using machine learning algorithms. The approach enabled seamless automation of pixel-wise analysis over multiple time periods and facilitated visualization of spatiotemporal trends in vegetation and urban expansion. The results derived from these methods offer a robust foundation for understanding land use transformations and their ecological implications in one of the fastest urbanizing regions of Bangladesh.

### *4.1 Data Selection*

The study area selected for this research is Dhaka District, one of the most urbanized and densely populated regions in Bangladesh. To ensure spatial accuracy and standardization, administrative boundary data was sourced from the Global Administrative Unit Layers (GAUL), a dataset developed by the Food and Agriculture Organization of the United Nations (FAO) in collaboration with the UN Cartographic Unit (UNCS). This dataset provides globally harmonized administrative boundaries with consistent coding and metadata. The technical metadata for the selected region, including administrative codes, country name, and spatial attributes such as shape area and length, is presented in ***Table 3***. The corresponding boundary map, illustrating the administrative extent of Dhaka District as used in this study, is shown in ***Figure 6***. This standardized geospatial input forms the foundation for subsequent image classification, index calculation, and land cover change analysis.

**Table 3** Technical metadata of Dhaka District based on the FAOUN GAUL 500m dataset, including administrative codes and spatial attributes.

| Name | Type | Description | Region of Interest |
|---|---|---|---|
| ADM0_CODE | Int | GAUL country code | 23 |
| ADM0_NAME | String | UN country name | Bangladesh |
| DISP_AREA | String | Unsettled territory: 'Yes' or 'No' | No |
| STATUS | String | Status of the country | Member State |
| Shape_Area | Double | Shape area | 0.130020260596 |
| Shape_Leng | Double | Shape length | 3.22574930953 |
| ADM1_CODE | Int | GAUL code of administrative units at first level | 577 |
| ADM1_NAME | String | Name of administrative units at first level | Dhaka |
| ADM2_CODE | Int | GAUL code of administrative units at second level | 5778 |
| ADM2_NAME | String | Name of administrative units at second level | Dhaka |
| EXP2_YEAR | Int | Expiry year of the administrative unit | 3000 |
| STR2_YEAR | Int | Creation year of the administrative unit | 1000 |

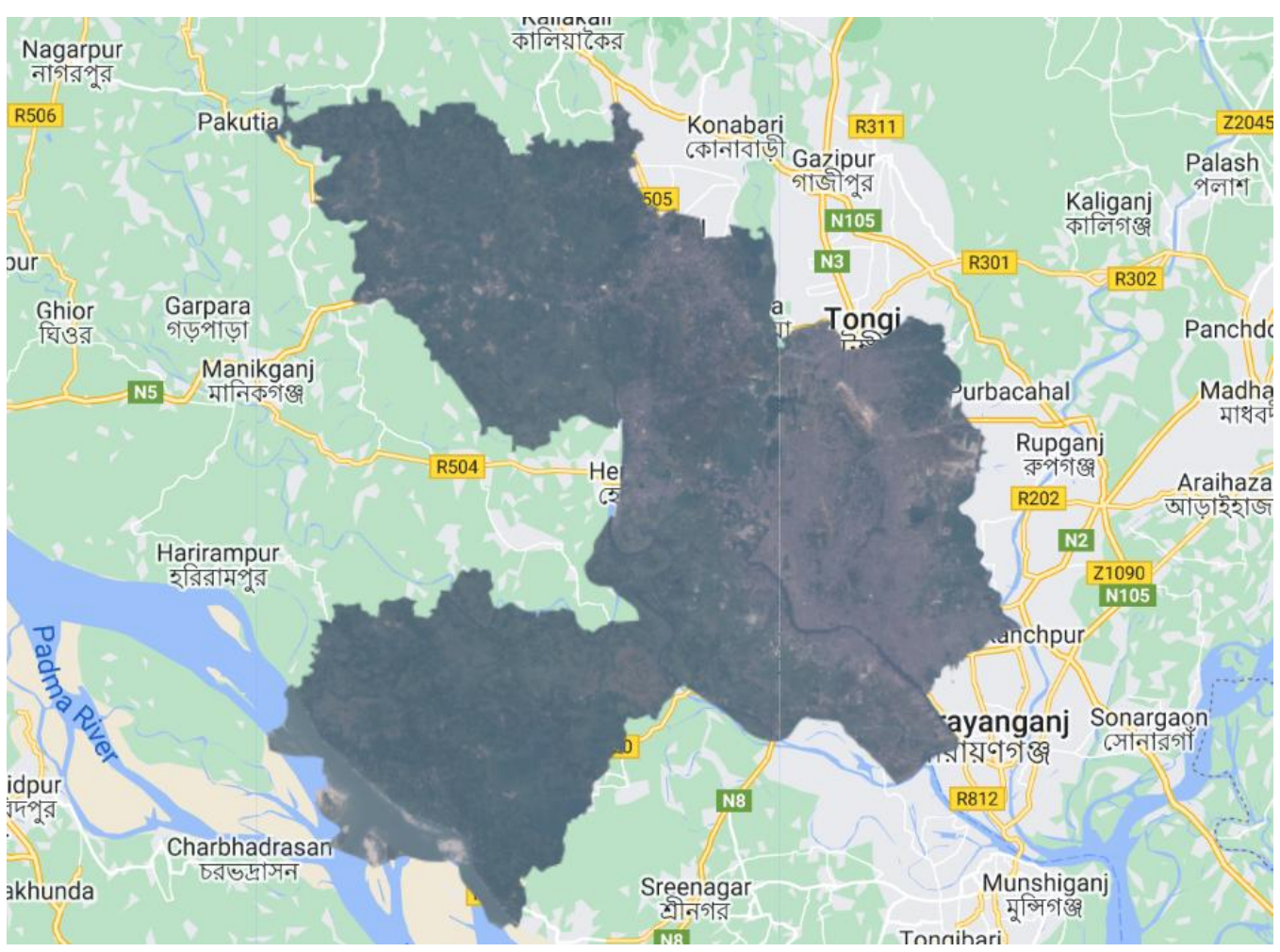

**Figure 6** Administrative boundary map of Dhaka District derived from the FAOUN GAUL dataset.

### *4.2 Multispectral Satellite Imagery and Image Preprocessing*

This study employs Sentinel-2 multispectral imagery, which offers high-resolution data across 13 spectral bands and supports a wide range of geospatial applications under the European Space Agency's Copernicus Land Monitoring initiative. Sentinel-2's data is particularly valuable for monitoring land cover, vegetation health, water bodies, and urban dynamics due to its high spatial, spectral, and temporal resolution.

#### *4.2.1 Cloud Removal in Multispectral Satellite Imagery*

Cloud contamination presents a major obstacle in multispectral image analysis, especially in tropical regions like Bangladesh where cloud cover is frequent and seasonally persistent. In Sentinel-2 Level 1C imagery, cloud coverage typically ranges from 10% to 30%, which can significantly hinder accurate land surface observation. To mitigate this, cloud masking techniques were applied to exclude cloud-affected pixels. Following this, a temporal median compositing approach was used, whereby images from a given period (e.g., January 1 to December 31, 2019) were processed to generate a cloud-free composite using only images with less than 30% cloud coverage.

In contrast, Sentinel-2 Level 2A imagery, which includes atmospheric correction and pre-applied cloud masks, required no additional cloud removal. These images were directly used for analysis due to their improved surface reflectance accuracy and cleaner pixel quality.

In this study, custom-trained machine learning land cover classification models were applied to cloud-free Sentinel-2 imagery for the years 2019, 2021, and 2023 to detect spatiotemporal changes. Additionally, a FAOUN-provided land cover product was used for the year 2021, as this was the most recent dataset publicly available for Dhaka District in the FAO database. The integration of both custom and external classifications provided a comparative basis to validate classification accuracy and consistency.

Through this multi-tiered preprocessing strategy, Sentinel-2 imagery was optimized for high-quality land cover classification, vegetation monitoring, and change detection, even in the presence of partial cloud contamination.

### *4.3 Training Dataset*

#### *4.3.1 Training Data: ESA WorldCover 10m 2021 Pretrained Dataset*

The ESA WorldCover 10m 2021 dataset, developed by the European Space Agency (ESA) in collaboration with leading research institutions, serves as a high-resolution global land cover product. It provides land cover classifications at a 10-meter spatial resolution using Sentinel-1 SAR and Sentinel-2 multispectral imagery. The dataset incorporates 12 land cover classes, including forests, urban areas, croplands, water bodies, and wetlands, and is built using machine learning-based classification models, making it suitable for both direct use and transfer learning.

This dataset offers several advantages for research and operational applications: (i) its global coverage supports large-scale environmental monitoring; (ii) it provides pretrained classification outputs, facilitating quick adaptation for model training and validation; (iii) it can serve as a benchmark for evaluating new algorithms; and (iv) it supports data fusion with higher-resolution or application-specific imagery to enhance classification accuracy.

#### *4.3.2 Usage in Land Cover Classification Models*

In this study, the ESA WorldCover dataset was leveraged to support supervised classification of satellite imagery for Dhaka District. It was used both as a training reference and as a benchmark dataset to evaluate the performance of custom-trained machine learning models. Its high spatial resolution and classification reliability allowed us to fine-tune classifiers while minimizing the need for extensive manual labeling. Additionally, the pretrained dataset supported model validation and comparison with externally generated land cover products, ensuring methodological robustness.

#### *4.3.3 Training Dataset for Supervised Classification*

To classify land cover in Dhaka District, labeled geospatial points were manually selected on Google Maps using visual interpretation of satellite images. These labeled points were categorized into four land cover classes: water body, vegetation, built-up/urban area, and soil/bare land. A total of 430 points were created and distributed across the district, as summarized in ***Table 4***. These training points were used to train and test supervised classification algorithms, including Decision Tree, K-Nearest Neighbor, and Random Forest classifiers.

**Table 4** Number of geospatially labeled points selected for training and testing supervised land cover classification models.

| **Class/Category** | **Number of labeled points/places** |
|---|---|
| Water Body | 101 |
| Vegetation | 153 |
| Buildup/Urban | 133 |
| Soil/Bare Land | 43 |

***Figure 7*** illustrates the spatial distribution of these labeled points across the district, while ***Figure 8*** shows a zoomed-in view of selected points for clarity. This stratified sampling approach ensured that training data adequately represented the spectral variability of different land cover types within the region, which is crucial for improving the generalizability and accuracy of classification models.

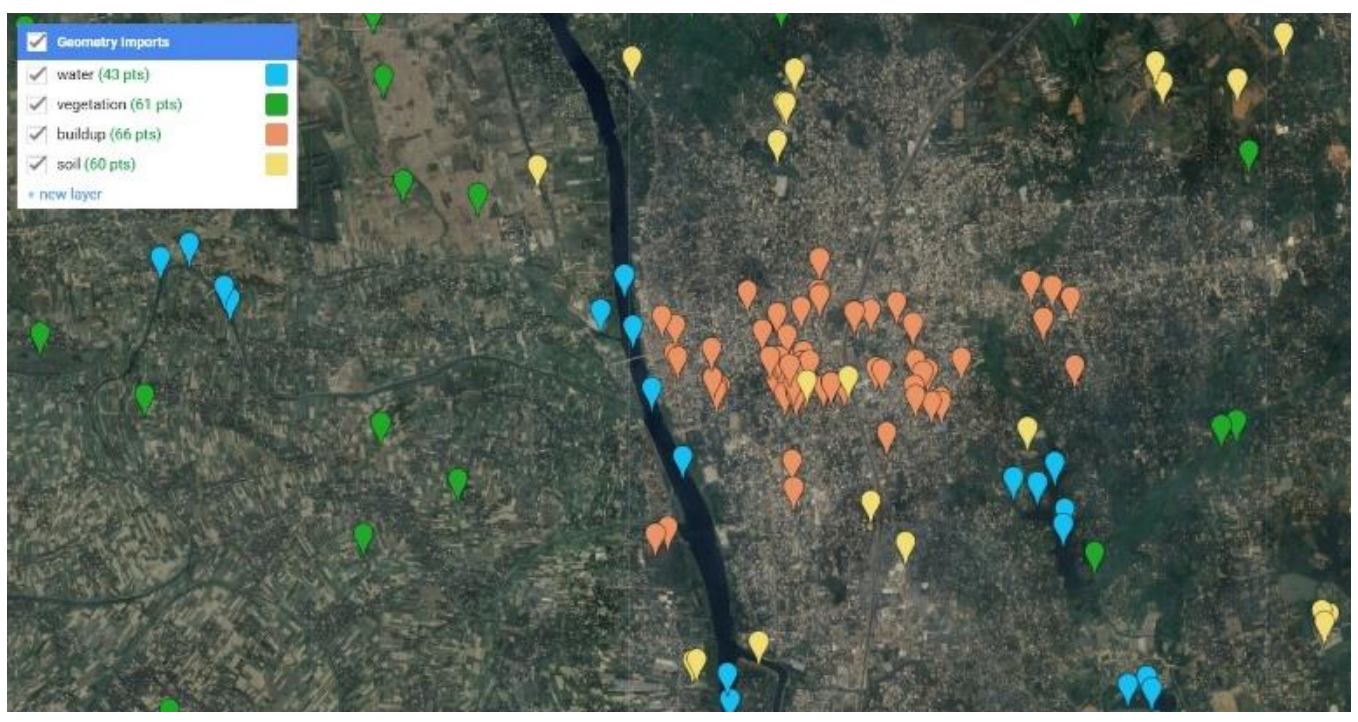

**Figure 7** Labeled training and testing points distributed across Dhaka District, representing four land cover classes as described in Table 4.

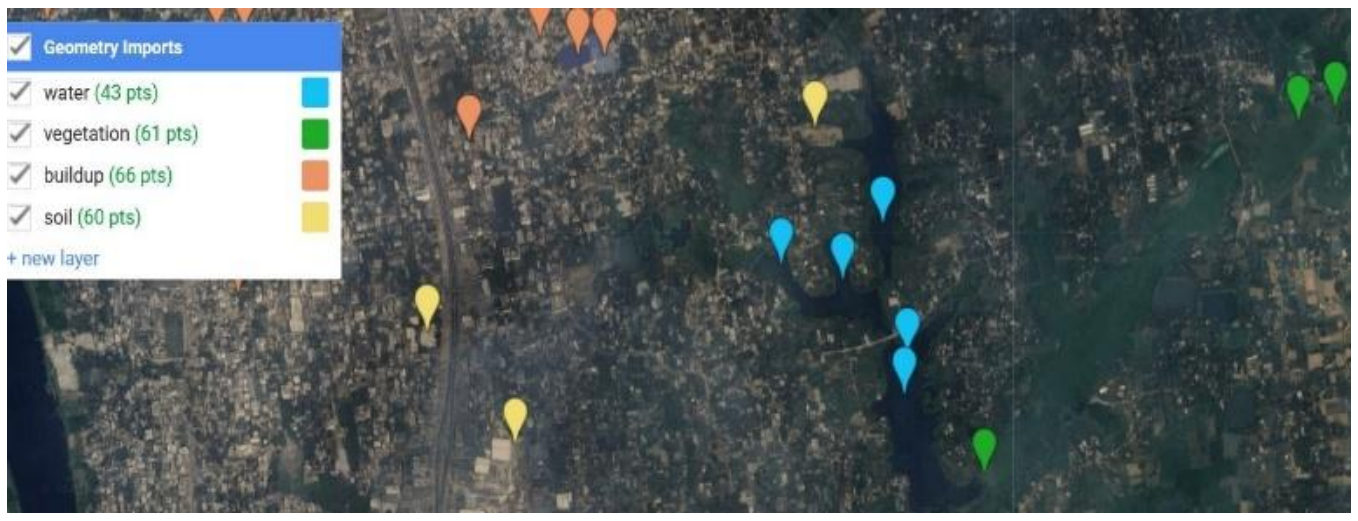

**Figure 8** Zoomed-in visualization of labeled training points for land cover classification (illustrative example).

### 4.4 Results and Discussion

#### *4.4.1 ESA WorldCover 10m 2021 (12 Classes)*

To evaluate land cover distribution across Dhaka District, we applied the ESA WorldCover 10m 2021 pretrained classification model, which was developed using Sentinel-1 and Sentinel-2 satellite data. Although the WorldCover model supports 12 global land cover classes, not all of them are present in Dhaka due to its specific geographic and climatic context. For instance, classes like snow and ice, moss and lichen, or mangroves are naturally absent in the region.

The resulting land cover classification for the year 2021 is illustrated in ***Figure 9***, clearly showing the spatial distribution of dominant classes such as cropland, tree cover, built-up areas, and water bodies. According to the model output, cropland is the most prominent class, covering approximately 632 square kilometers, followed by tree cover (413 km²), and built-up areas (239 km²). These statistics are detailed in ***Table 5***, which summarizes the area covered by each land class.

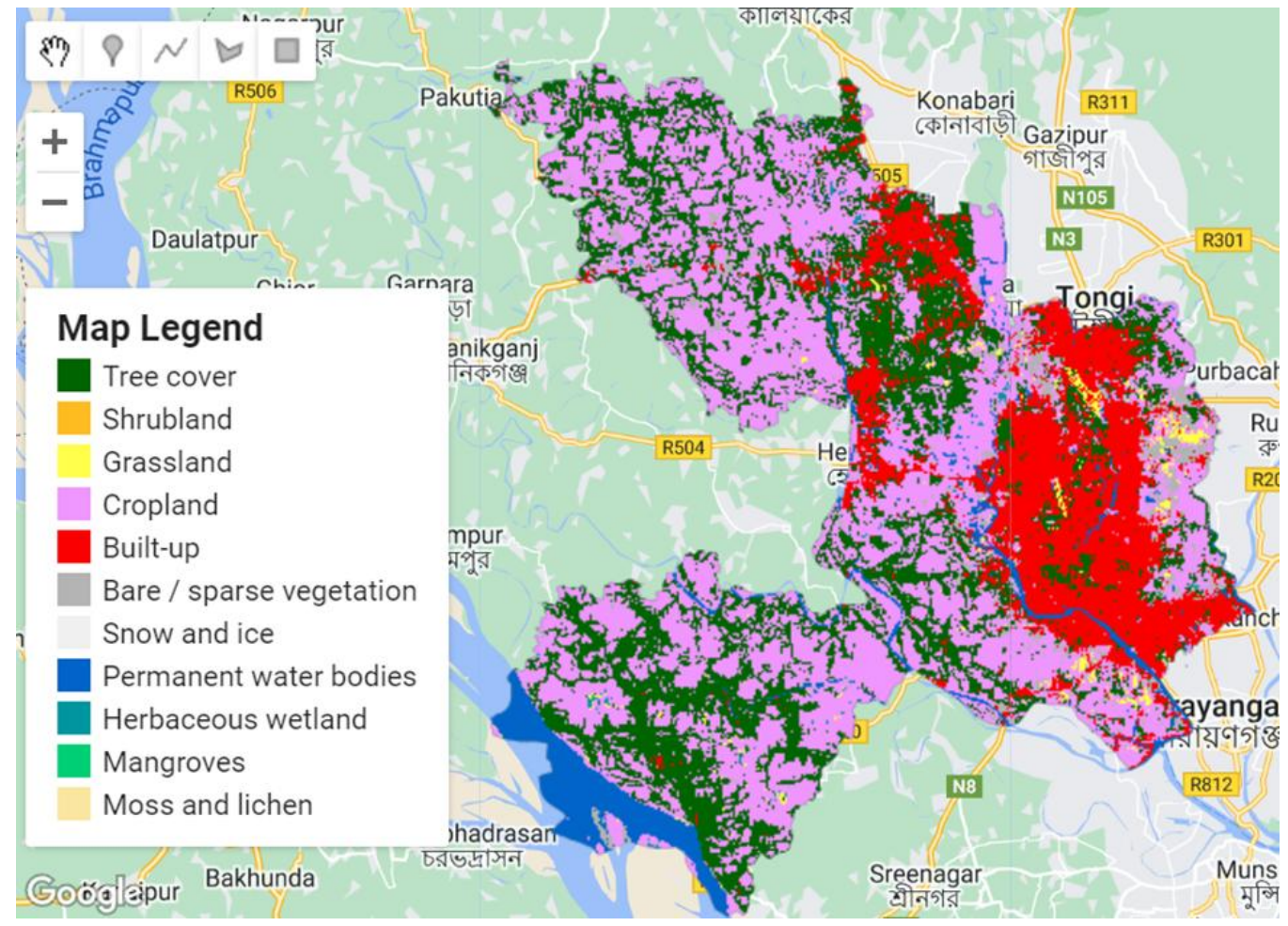

**Figure 9** ESA WorldCover 10m 2021-based land cover classification map of Dhaka District showing spatial distribution across major classes.

**Table 5** Land cover class-wise area distribution (in square kilometers) for Dhaka District in 2021, derived from ESA WorldCover dataset.

| Land Classes | Area (Square KM) |
|---|---|
| Bare / sparse vegetation | 75 |
| Built-up | 239 |
| Cropland | 632 |
| Grassland | 22 |
| Herbaceous wetland | 7 |
| Permanent water bodies | 79 |
| Tree cover | 413 |
| | **Total 1467 Sq. km** |

The total land area classified by the model is 1,467 square kilometers, which is remarkably close to the official administrative area of Dhaka District (1,471 km²), indicating the high accuracy of the ESA WorldCover dataset. A proportional overview of the land classes is also presented in the 3D pie chart shown in ***Figure 10***, offering a visual interpretation of class-wise land coverage distribution.

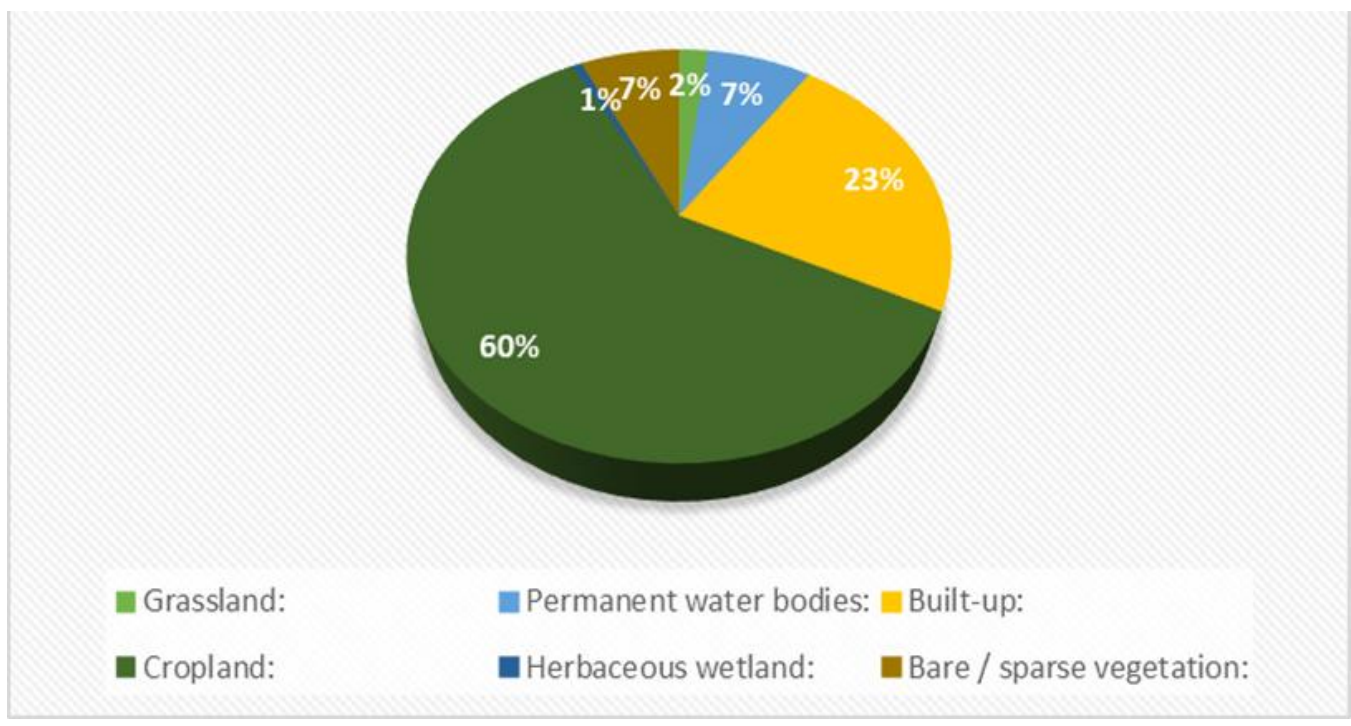

**Figure 10** 3D pie chart illustrating the proportional land cover distribution in Dhaka District based on the 2021 WorldCover classification.

### *4.4.2 Landcover Classification and Change Detection*

To assess the dynamic transformation of land cover in Dhaka District over time, a set of supervised machine learning models—Decision Tree, Random Forest, and K-Nearest Neighbors (KNN)—were applied separately on multispectral satellite imagery from the years 2019 and 2024. Each classifier was evaluated using standard performance metrics including overall accuracy, F1-score, kappa coefficient, consumer's accuracy, and producer's accuracy, allowing for a detailed comparison of model effectiveness.

The performance comparison for the 2019 classification is summarized in ***Table 6***. Among the three models, Random Forest consistently outperformed others, achieving the highest F1-scores across most land classes and the strongest kappa coefficient of 0.8362. Notably, water and vegetation classes were classified with very high accuracy in all models, while bare land presented challenges, as reflected in its lower F1-score values. This comparative analysis demonstrates that Random Forest is the most reliable model for land cover classification in this context, due to its robustness in handling nonlinear boundaries and noise in the training data.

**Table 6** Performance comparison of Decision Tree, Random Forest, and KNN models for 2019 land cover classification in Dhaka District.

| **Metric** | **Classes** | **Decision Tree** | **Random Forest** | **KNN, k= 4** |
|---|---|---|---|---|
| Confusion Matrix | Class 0 = Urban | [26,3, 0,1] | [28,1, 0,1] | [24,6, 0,0] |
| | Class 1 = Bare | [4,5, 0,1] | [4,5, 0,1] | [2,7, 0,1] |
| | Class 2 = Water | [0,0, 28,0] | [0,0, 27,1] | [0,0, 27,1] |
| | Class 3 = Vegetation | [1,1, 1,32] | [2,1, 1,31] | [3,0, 1,31] |
| Accuracy | Combined | 0.8835 | 0.8835 | 0.8641 |
| F1-Score | Class 0 = Urban | 0.8525 | 0.8750 | 0.8136 |
| | Class 1 = Bare | 0.5263 | 0.5882 | 0.6087 |
| | Class 2 = Water | 0.9825 | 0.9643 | 0.9643 |
| | Class 3 = Vegetation | 0.9275 | 0.8986 | 0.9118 |

| Kappa | Combined | 0.8371 | 0.8362 | 0.8120 |
|---|---|---|---|---|
| Consumer's Accuracy | Class 0 = Urban | 0.8387 | 0.8235 | 0.8276 |
| | Class 1 = Bare | 0.5556 | 0.7143 | 0.5385 |
| | Class 2 = Water | 0.9655 | 0.9643 | 0.9643 |
| | Class 3 = Vegetation | 0.9412 | 0.9118 | 0.9394 |
| Producer's Accuracy | Class 0 = Urban | 0.8667 | 0.9333 | 0.8000 |
| | Class 1 = Bare | 0.5000 | 0.5000 | 0.7000 |
| | Class 2 = Water | 1.0000 | 0.9643 | 0.9643 |
| | Class 3 = Vegetation | 0.9143 | 0.8857 | 0.8857 |

The output of the Random Forest classification for 2019 is shown in ***Figure 11***, where spatially distinct land categories, urban areas, vegetation, water bodies, and bare soil, are visibly mapped across the Dhaka District. This classification was trained using 430 manually labeled points, which were evenly distributed across all classes to ensure representativeness and reduce model bias.

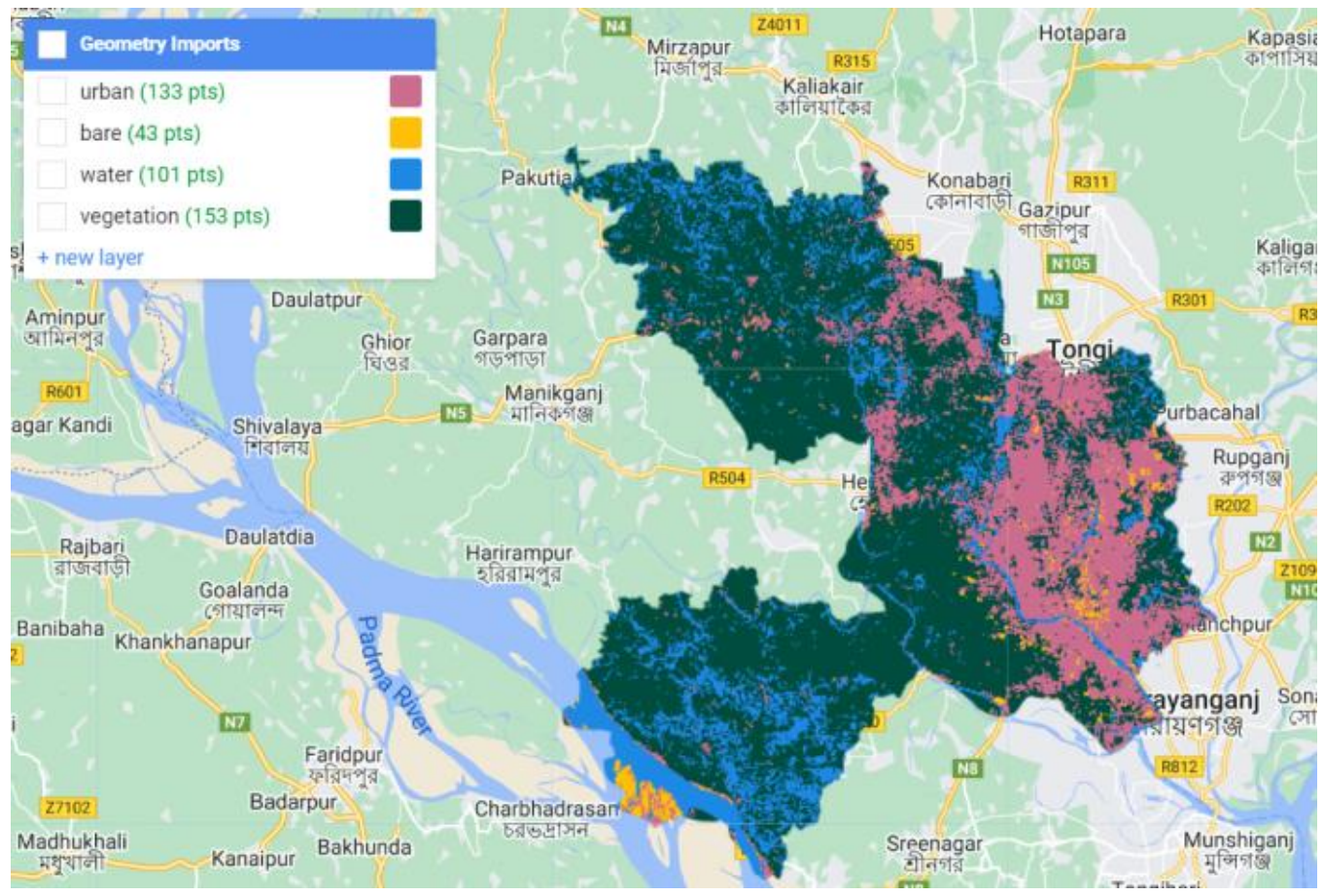

**Figure 11** Land cover classification map of Dhaka District in 2019 using the Random Forest model.

Subsequently, the same model and methodology were applied to satellite imagery from 2024, with the resulting land classification presented in ***Figure 12***. Visual comparison of the 2019 and 2024 classification maps reveals notable shifts in land cover, particularly a significant expansion in urban areas and a corresponding reduction in both vegetation and water bodies.

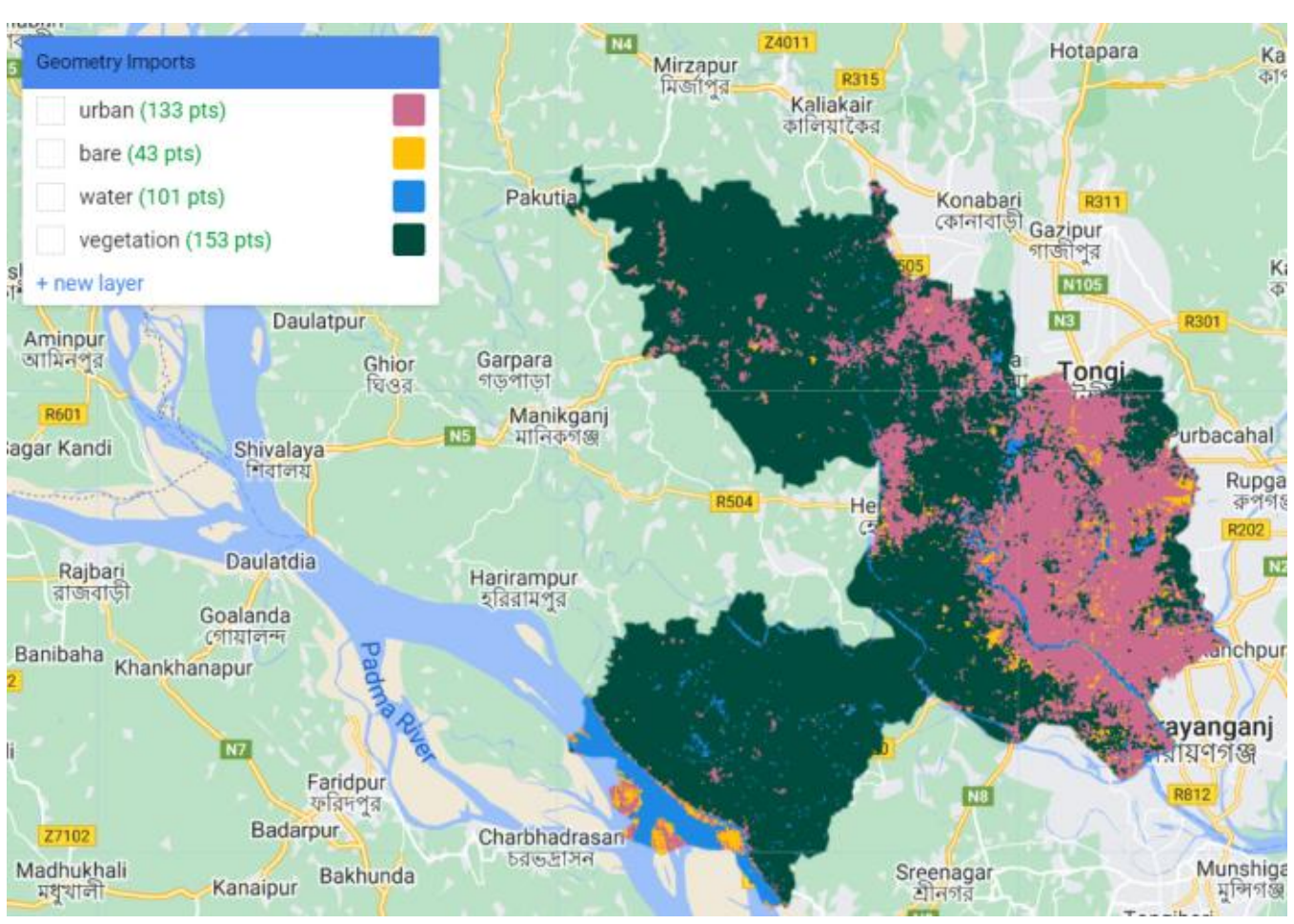

**Figure 12** Land cover classification map of Dhaka District in 2024 using the Random Forest model.

A side-by-side quantitative comparison of land cover areas between 2019 and 2024 is illustrated in ***Figure 13***, and the corresponding numerical values are documented in ***Table 7***. The data reveals a 53 square kilometer increase in urban area, signaling rapid urban expansion. Simultaneously, vegetation decreased by 22 km², and water bodies shrank by 46 km², while bare land experienced a modest 15 km² increase. These changes indicate a shift in land use patterns, likely driven by infrastructure development, urban encroachment, and natural resource depletion.

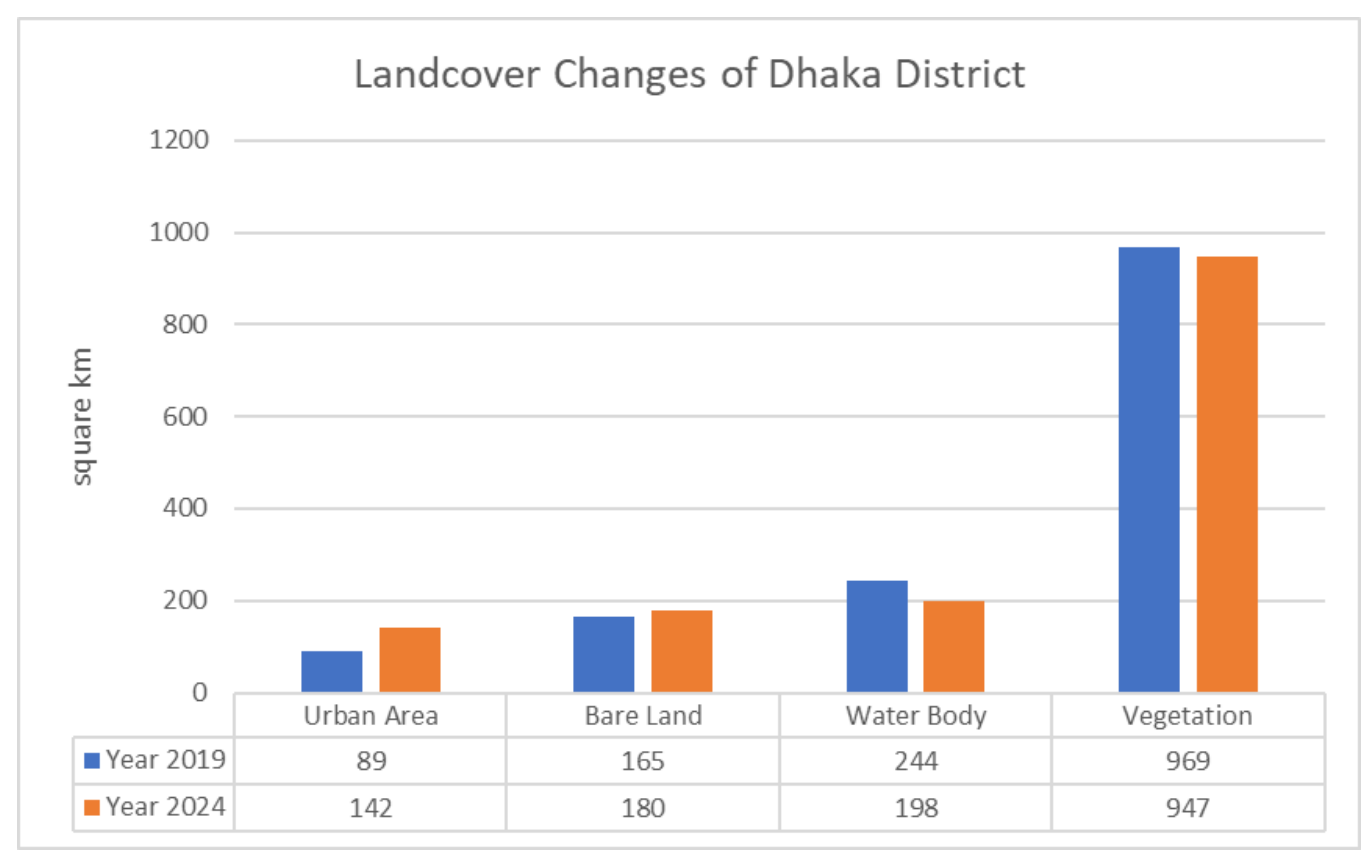

**Figure 13** Comparative bar chart showing land cover area changes between 2019 and 2024.

**Table 7** Summary of land cover changes (in square kilometers) between 2019 and 2024 in Dhaka District.

| Classes | Dhaka in 2019 Landcover (Sq. km) | Dhaka in 2024 Landcover (Sq. km) | Differences (Sq. km) | Remarks |
|---|---|---|---|---|
| Urban Area | 89 | 142 | 53 | Increased |
| Bare Land | 165 | 180 | 15 | Increased |
| Water Body | 244 | 198 | -46 | Decreased |
| Vegetation | 969 | 947 | -22 | Decreased |

A detailed spatial visualization of land cover changes—specifically the locations where transitions occurred from one land class to another—is shown in ***Figure 14***. The red-highlighted areas represent all pixels that experienced class change between 2019 and 2024, underscoring the widespread nature of landscape transformation across Dhaka District.

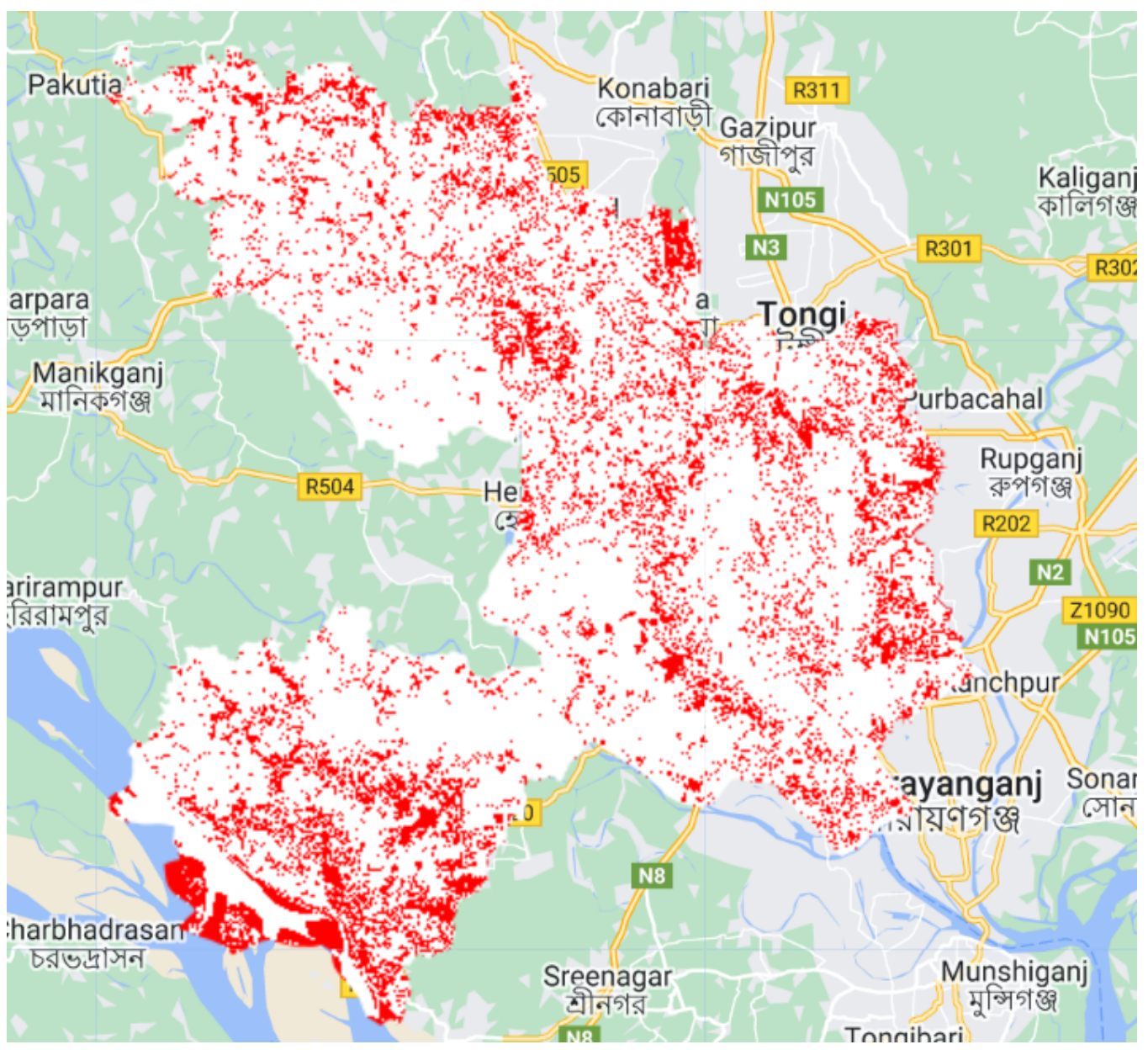

**Figure 14** Spatial distribution of all land cover class changes between 2019 and 2024.

To understand the interclass transitions, ***Table 8*** presents a matrix of land cover conversions from 2019 to 2024. For instance, 62 km² of land that was classified as vegetation in 2019 was converted into urban area by 2024, while an additional 12 km² of bare land also transitioned to urban use. These conversions are visualized in ***Figure 15***, which highlights the specific areas where vegetation was replaced by built-up infrastructure, providing powerful spatial evidence of the ongoing urbanization process.

**Table 8** Interclass transition matrix indicating specific land cover conversions from 2019 to 2024.

| | | 2024 | | | |
|---|---|---|---|---|---|
| | | Urban | Bare land | Waterbodies | Vegetation |
| **2019** | Urban | - | 8 | 5 | 13 |
| | Bare Land | 12 | - | 4 | 2 |
| | Waterbodies | 8 | 6 | - | 149 |
| | Vegetation | 62 | 17 | 3 | - |

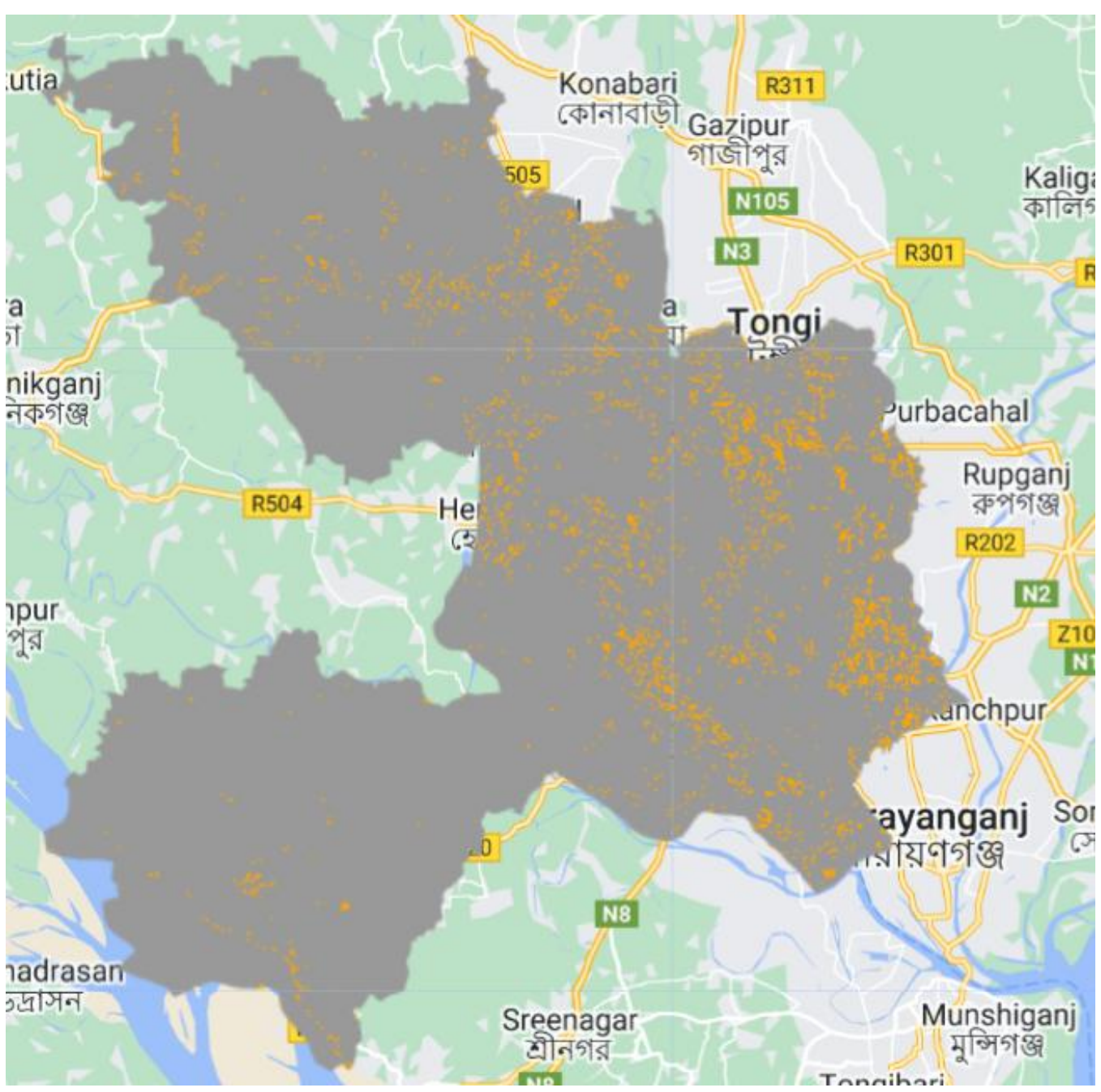

**Figure 15** Map showing detected changes from vegetation to built-up land between 2019 and 2024.

This integrated spatiotemporal analysis provides critical insight into the urban expansion patterns and ecological transformation occurring in Dhaka District. The alarming decline in vegetation and water bodies underscores the need for sustainable urban planning, policy interventions, and environmental monitoring to mitigate the long-term impacts of unregulated development.

#### *4.4.3 Land Classification Using Multispectral Band Difference Ratio*

As described previously, different land surface materials reflect electromagnetic radiation differently across spectral bands such as red, green, blue, near-infrared (NIR), and shortwave infrared (SWIR). These variations can be effectively captured using band ratio-based indices, allowing for efficient land cover classification without relying on machine learning.

We utilized three widely accepted spectral indices: the Normalized Difference Vegetation Index (NDVI), the Normalized Difference Built-up Index (NDBI), and the Normalized Difference Water Index (NDWI) to detect vegetation, urban, and water body areas respectively. These indices were computed for both 2019 and 2024 over the Dhaka District using Sentinel-2 multispectral imagery.

The NDVI-based vegetation cover is illustrated in ***Figure 16***, comparing the vegetation extent in 2019 (left) and 2024 (right). A visible reduction in green-covered areas is observed, reflecting a trend of declining vegetation due to urban expansion and possible land conversion.

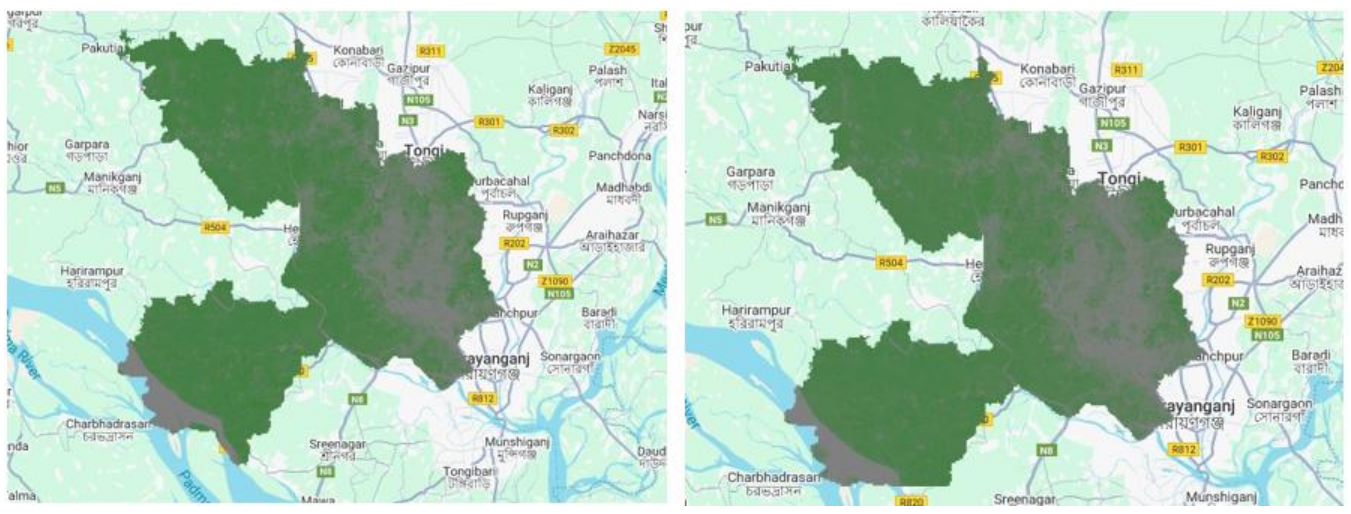

**Figure 16** NDVI-based vegetation cover of Dhaka District for 2019 (left) and 2024 (right), showing a visible decrease in green areas.

Similarly, ***Figure 17*** presents the results of NDWI-based water body detection for both years. The images clearly show a loss of water bodies between 2019 and 2024, particularly in the southern and western parts of the district. This decline can be attributed to land reclamation, drainage blockage, or anthropogenic encroachments.

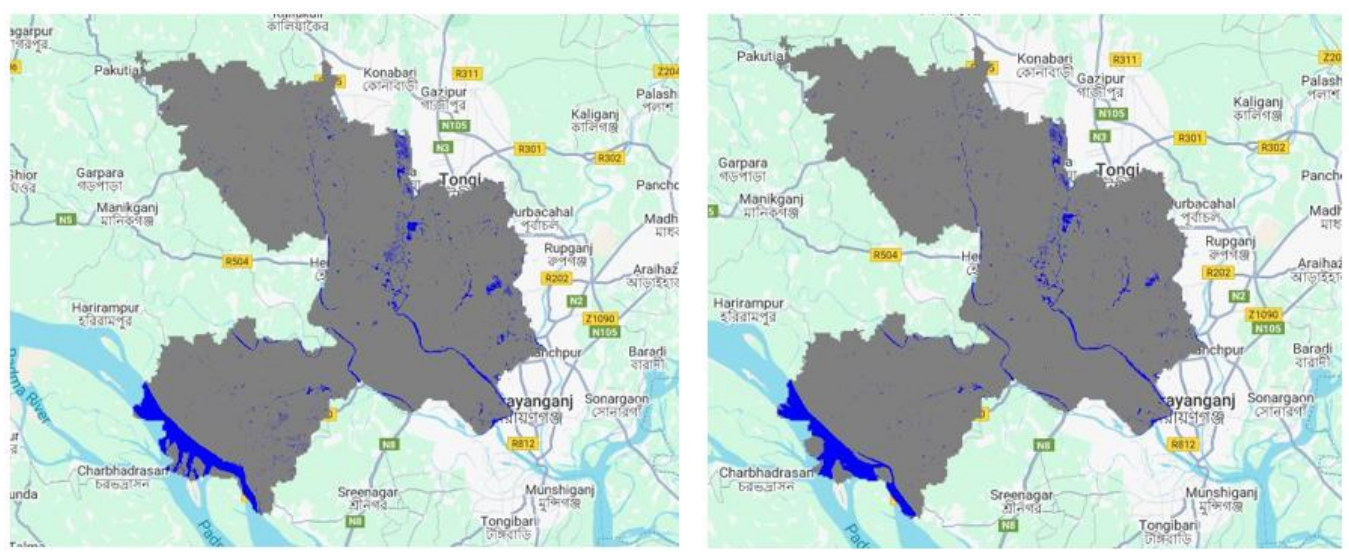

**Figure 17** NDWI-based detection of water bodies in 2019 (left) and 2024 (right), revealing substantial water loss.

The NDBI index, shown in ***Figure 18***, highlights built-up areas. The 2024 map (right) indicates a significant spread of urban infrastructure, especially in and around central Dhaka and the northern fringe, compared to the 2019 map (left). This expansion aligns with the city's ongoing population growth and development activities.

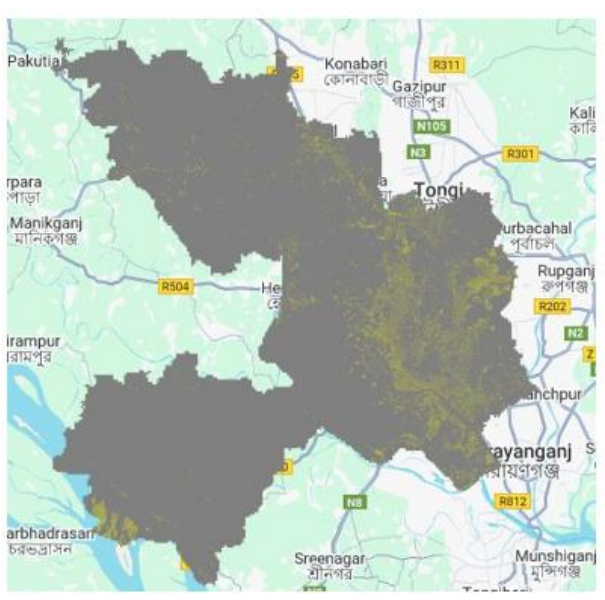

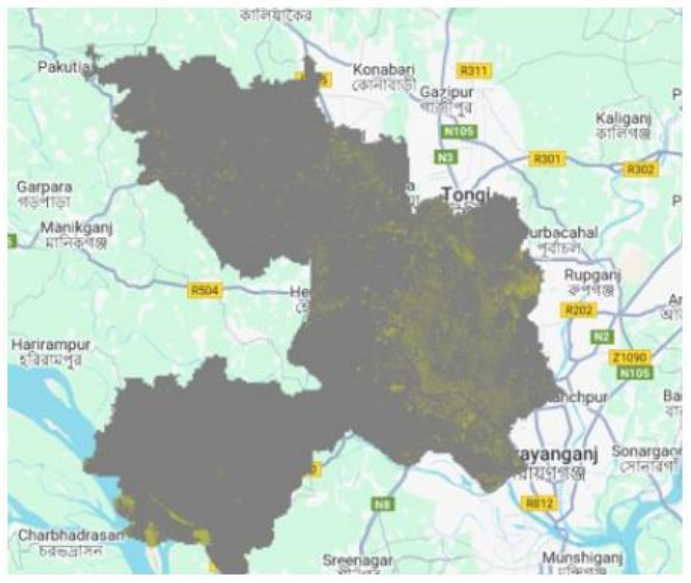

**Figure 18** NDBI-based detection of built-up areas in 2019 (left) and 2024 (right), highlighting urban sprawl.

A quantitative summary of the detected land cover changes from these indices is provided in ***Table 9***. According to the results:

- Vegetation declined from 945 km² in 2019 to 865 km² in 2024—a net loss of 80 km², representing an 8.46% decrease.
- Urban area expanded from 278 km² to 302 km²—an increase of 24 km², approximately 8.63% growth.
- Water bodies reduced from 90 km² to 83 km², marking a 7.77% decline.

**Table 9** Summary of land cover changes detected using NDVI, NDWI, and NDBI indices, quantifying percentage and area changes for each class.

| | Area (sq. km) | | | |
|---|---|---|---|---|
| **Class** | **2019** | **2023** | **Change (sq. km)** | **Change (%)** |
| Vegetation | 945 | 865 | -80 | -8.46 |
| Urban | 278 | 302 | +24 | +8.63 |
| Water | 90 | 83 | -7 | -7.77 |

These findings align with the outputs of supervised machine learning models discussed earlier but with slight discrepancies. It is important to note the methodological differences:

- The normalized difference index approach focuses only on three land classes (vegetation, water, urban), excluding bare land, which may have been partially absorbed into the urban or vegetation categories.
- Classification depends heavily on threshold values, and slight modifications to these thresholds can influence class boundaries and area estimations.
- Unlike machine learning methods, index-based approaches do not utilize contextual spatial relationships or multi-class segmentation, which limits their granularity.

Despite these differences, the trend remains consistent: urban land is increasing at the expense of vegetation and water bodies, a pattern indicative of rapid urbanization and ecological stress.

### *4.5 Research Answers (RA) to the Research Questions*

***RA1:*** The major land cover classes identified in Dhaka District include urban/built-up areas, vegetation, water bodies, and bare land. Analysis of Sentinel-2 imagery from 2019 and 2024 revealed significant changes over the five-year period. Urban areas increased substantially (by approximately 53 km²), while vegetation and water bodies showed marked declines—22 km² and 46 km² respectively (***Table 7***, ***Figures 11-14***). These trends reflect rapid urbanization and ecological transformation in the region.

***RA2:*** Machine learning models demonstrated high effectiveness in classifying land cover types using multispectral data. Among the models tested, Random Forest outperformed both Decision Tree and KNN, achieving the highest overall accuracy (88.35%), F1-scores (up to 0.89 for vegetation), and kappa coefficient (0.8362) (***Table 6***). It showed strong precision in classifying vegetation and water, with some limitations in bare land classification due to spectral similarity with other classes.

***RA3:*** Spectral index-based methods (NDVI for vegetation, NDWI for water, and NDBI for built-up areas) provided reliable but less detailed classifications than machine learning approaches. The results from index-based classification closely mirrored the machine learning outcomes, confirming a vegetation decrease of 80 km² (-8.46%), urban increase of 24 km² (+8.63%), and water body loss of 7 km² (-7.77%) between 2019 and 2024 (***Table 9***, ***Figures 16–18***). However, the index-based method classified only three classes and had limitations in distinguishing bare land, which could lead to classification ambiguity.

***RA4:*** The study revealed clear spatiotemporal trends in land cover change. Urban areas expanded primarily in the north and central parts of Dhaka, often replacing previously vegetated zones (***Figure 15***). Vegetation loss was prominent across peripheral and agricultural zones, while water body reduction was observed in the southern and southwestern regions. These changes indicate strong anthropogenic pressure, urban encroachment, and environmental degradation across the district over just five years.

***RA5:*** The findings provide actionable insights for policymakers and urban planners. The alarming decrease in vegetation and water bodies underscores the urgency for sustainable land management, urban growth regulation, and wetland conservation strategies. By integrating machine learning and remote sensing technologies, authorities can monitor urban sprawl, protect ecological resources, and guide infrastructure development in harmony with environmental preservation. Regular land use assessments should be institutionalized for proactive and informed decision-making.

The experimental results clearly demonstrate significant land cover changes in Dhaka, marked by rapid urban expansion and notable declines in vegetation and water bodies. Among the classification approaches, Random Forest yielded the most accurate results, while spectral indices effectively supported change detection. These findings highlight the power of integrating machine learning and remote sensing for monitoring urbanization and underscore the need for sustainable planning and environmental protection.

## 5. Conclusion

This study explored land cover classification and vegetation dynamics in Dhaka District using a combination of remote sensing, spectral indices (NDVI, NDWI, NDBI), and machine learning algorithms including Random Forest, Decision Tree, and KNN. The research focused on detecting spatiotemporal changes from 2019 to 2024, revealing a substantial increase in urban areas alongside a decline in vegetation and water bodies, as evidenced by both supervised classification and normalized difference index methods. Among the models, Random Forest consistently outperformed others, demonstrating high accuracy and robustness in land classification tasks. The integration of ESA WorldCover data and Sentinel-2 imagery enabled a detailed and scalable analysis, contributing valuable methodological insights to the field of geospatial analysis. From a policy perspective, the findings emphasize the urgency of implementing sustainable urban planning and environmental conservation strategies, particularly in rapidly urbanizing regions like Dhaka. By identifying key areas of ecological loss and urban expansion, this research offers a scientific basis for targeted interventions and long-term urban resilience planning. Overall, the study demonstrates the effectiveness of combining machine learning with satellite-based monitoring and contributes both technically and practically to the advancement of land use analysis and environmental decision-making.

Future research can explore the integration of higher-resolution commercial satellite imagery and deep learning models such as convolutional neural networks (CNNs) to improve classification accuracy and detect finer-scale urban dynamics. Additionally, incorporating socio-economic and climate data could enhance the understanding of drivers behind land cover change, supporting more comprehensive urban and environmental planning.

**Availability of Data and Coding:** The datasets generated and analyzed during the current study, as well as the code scripts used for processing and classification in Google Earth Engine, are available from the corresponding author upon reasonable request.

**Conflict of Interest:** The authors declare that they have no conflict of interest regarding the publication of this manuscript.

**Funding:** This research did not receive any specific grant from funding agencies in the public, commercial, or not-for-profit sectors.

**Clinical Trial Number**: Not Applicable

**Ethics, Consent to Participate, and Consent to Publish Declarations**: Not Applicable

**Author’s Contribution**:

Muhammad Masud Tarek: Conceptualization, Methodology, Remote Sensing Analysis, Writing – Original Draft Preparation, Supervision.

Md. Alamgir Hossain: Spectral Index Computation, Validation, Result Interpretation, Literature Review, Writing – Review & Editing.

Md. Samiul Islam: Data Collection, Image Preprocessing, Land Cover Classification, Visualization, Writing – Review & Editing.

Muntasir Hasan Kanchan: Coding in Google Earth Engine, Machine Learning Implementation, Accuracy Assessment, Data Curation, Manuscript Formatting.

"All authors have read, understood, and have complied as applicable with the statement on "Ethical responsibilities of Authors"